%% file: emnlp2022.tex
\pdfoutput=1

\documentclass[11pt]{article}

\usepackage{EMNLP2022}

\usepackage{times}
\usepackage{latexsym}
\usepackage{graphicx}
\usepackage{multirow}
\usepackage{array}

\usepackage[T1]{fontenc}

\usepackage[utf8]{inputenc}

\usepackage{microtype}

\usepackage{inconsolata}

\title{Do Language Models Understand Measurements?}

\author{
    Sungjin Park \quad Seungwoo Ryu \quad Edward Choi \\
    KAIST\\
    \texttt{\{zxznm, swryu, edwardchoi\}@kaist.ac.kr}
}

\begin{document}
\maketitle
\begin{abstract}
\input{00Abstract}

\end{abstract}

\section{Introduction}
\label{sec:Introduction}
\input{01Introduction}


\section{Measuring Skill Test}
\label{sec:Probing}
\input{02Probing}

\section{Experiments}
\label{sec:Experiments}
\input{03Experiments}

\section{Results and Analysis}
\label{sec:Results}
\input{04Results}


\section{Related Works}
\label{sec:RelatedWorks}

\input{08RelatedWorks}

\section{Conclusion}
\label{sec:Conclusion}
\input{05Conclusion}

\section*{Acknowledgements}
Research supported with Cloud TPUs from Google's TPU Research Cloud (TRC).
This work was supported by Institute of Information \& Communications Technology Planning \& Evaluation (IITP) grant (No.2019-0-00075),
the Korea Health Industry Development Institute (KHIDI) grant (No.HI21C1138),
and the Korea Medical Device Development Fund grant (Project No.: 1711138160, KMDF\_PR\_20200901\_0097), funded by the Korea government (MSIT, MOTIE, MOHW, MFDS).

\section*{Limitations}
\input{07Limitation}

\bibliography{anthology, custom}
\bibliographystyle{acl_natbib}

\clearpage
\appendix

\input{06Appendix}

\end{document}

%% file: 00Abstract.tex
Recent success of pre-trained language models (PLMs) has stimulated interest in their ability to understand and work with numbers. 
Yet, the numerical reasoning over measurements has not been formally studied despite their importance.
In this study, we show that PLMs lack the capability required for reasoning over measurements.
Furthermore, we find that a language model trained on a measurement-rich corpus shows better performance on understanding measurements. 
We propose a simple embedding strategy to better distinguish between numbers and units, which leads to a significant improvement in the probing tasks.

%% file: 01Introduction.tex
The success of pre-trained language models (PLMs) has led to more research on their ability to understand commonsense.
In this context, numerical reasoning over text (NRoT) is a NLP model's ability to interpret and work with numbers in either digit or word form~\cite{spithourakis-riedel-2018-numeracy}.
Recent studies on NRoT test PLMs to answer questions on numeracy~\cite{wallace-etal-2019-nlp}, scalar magnitude comparison~\cite{zhang-etal-2020-language-embeddings}, numerical facts~\cite{lin-etal-2020-birds}, and math word problems~\cite{wu-etal-2021-math}.

Despite these efforts, existing works lack an analysis of the forms in which numbers appear. 
In particular, we focus on the case where numbers appear as a measurement in the context. 
In most scientific articles, measurements are an integral part of the context for capturing its appropriate meaning. 
For example, the two sentences "40g of Aspirin is lethal" and "40mg of Aspirin is lethal" contain the same words except for the unit of measurement (UoM), but the second sentence is incorrect because of the UoM. 

In this work, we examine the \textit{measuring skill} of PLMs: the ability to understand the system of measurement and perform numerical reasoning over measurements.
We design three measuring skill tests (MSTs) and study how many measuring skills can be acquired.
Specifically, \textsc{unit conversion}, \textsc{reference range detection}, and \textsc{measurement comparison} require understanding of the system of measurement, the normal range of the biomedical entity, and the ability to combine knowledge about the system of measurement and NRoT, respectively.
Table~\ref{tab:example} shows an example of each of the measuring skill tests.

MST results showed that the models struggled to find the largest (or smallest) value on the list of measurements and convert the measurement to another unit, while they performed well on other tests. 
Compared to other PLMs, BioBERT~\cite{lee2020biobert} showed superior performance on \textsc{unit conversion} and \textsc{reference range detection}, which implies that pre-training with measurement-rich text helps the model understand the system of measurement.
Finally, we speculate that the lack of skills to distinguish numbers, units, and other words in the context makes the models fail in some MSTs.
To mitigate this, we introduce \textit{scale embedding}, which provides the model with the information regarding the position and scale of the numbers in the input text.
We show that scale embedding significantly improves the MST performance of all PLMs.

\input{Table/01samples}


%% file: Table/01samples.tex
\begin{table*}[t]
    \centering
    \resizebox{2.0\columnwidth}{!}{%
    \begin{tabular}{l|ll}
        \hline 
        \textbf{Task} & \textbf{Example} & \textbf{Answer Candidates}\\
        \hline 
        \textsc{Comparison} & 1.59mg is $\mathsf{[MASK]}$ than 3.8g & larger, \underline{smaller}\\
        \textsc{Argmin/max} & $\mathsf{[MASK]}$ value among 0.5mg, 3.4g, 2.8mg is 0.5mg & largest, \underline{smallest}, middle \\
        \textsc{Sorting} & sort 0.53mg, 32.54g, 2.8mg in $\mathsf{[MASK]}$ order is 0.53mg, 32.54g, 2.8mg & increasing, decreasing, \underline{random}\\
        \textsc{Unit Conversion} & 3.5g and 3500mg are $\mathsf{[MASK]}$ value & \underline{same}, different \\ 
        \textsc{Reference Range Detection} & 85mg/dL of Glucose is $\mathsf{[MASK]}$ & \underline{normal}, abnormal \\
        \hline 
        
    \end{tabular}
    }
    \caption{Examples of measuring skill tests (MSTs). We underline the correct answer for each example.}
    \label{tab:example}
\end{table*}

%% file: 02Probing.tex
\input{Table/01_2_templates}


In this section, we describe three MSTs to carefully study the ability of PLMs to understand the system of measurement and perform numerical reasoning over the measurements. 
%


\subsection{Unit Conversion}
\label{subsec:unit_conversion}
This task requires the model to decide whether the two measurements represent the same quantity.
For example, the model might correctly predict \texttt{[MASK]} in a sentence, such as \textit{"3.5g and 3500mg are \texttt{[MASK]} value"} to be filled with \textit{same} if it understands the conversion of units correctly.
In general, it is a convention to combine the unit (e.g., liter, meter) and its prefix (e.g., kilo, milli) to represent the numerical value of the measurement within a range $[10^{-3}, 10^{3})$.
Therefore, various unit prefixes can appear in a single passage, even if the units are the same.
To handle this, \textsc{unit conversion} is essential for complex reasoning over measurements.
To succeed in \textsc{unit conversion}, we expect the model to handle the unit and numerical value jointly, based on an understanding of the system of measurement.

\subsection{Reference Range Detection}
\label{subsec:ref_range}
Given a biomedical entity and measurement, this task requires a model to predict whether the measurement falls within the reference range.
Knowledge of the biomedical entity plays a crucial role in understanding measurements, since the unit is determined by the biomedical entity.
For example, we measure the hemoglobin level in g/dL.
In addition to understanding UoMs, PLMs must rely on domain knowledge embedded in their parameters to solve this task, as context alone does not provide sufficient clues as to what the reference range is for the given biomedical entity.

\subsection{Measurement Comparison}
\label{subsec:measurement_comp}
Given two measurements (or a series of \textit{n} measurements), the task is to predict the correct relationship between them.
We created the synthetic dataset following other well-known NRoT tasks.
Here, we consider three numerical reasoning tasks: \textsc{Comparison}~\cite{talmor-etal-2020-olmpics}, \textsc{ArgMin/Max}~\cite{wallace-etal-2019-nlp}, and \textsc{Sorting}~\cite{pal-baral-2021-investigating-numeracy}, all requiring the model to compare numbers.
Note that each measurement in this task can have a different unit prefix.
For example, the sample \textit{"1.59mg is \texttt{[MASK]} than 3.8g"} containing two different units "mg" and "g" appears in the \textsc{Comparison} dataset.
This task assesses the model's ability to combine an understanding of measurements and numerical reasoning skills.

%% file: Table/01_2_templates.tex
\begin{table}[t]
    \centering
    \resizebox{1.0\columnwidth}{!}{%
    \begin{tabular}{l|l}
        \hline 
        \textbf{Task} & \textbf{Template}\\
        \hline 
        \textsc{Comparison} & $\mathsf{[M]}$ is $\mathsf{[MASK]}$ than $\mathsf{[M]}$\\
        \textsc{Argmin/max} & $\mathsf{[MASK]}$ value among $\mathsf{[LoM]}$ is $\mathsf{[M]}$ \\
        \textsc{Sorting} & sort $\mathsf{[LoM]}$ in $\mathsf{[MASK]}$ order is $\mathsf{[LoM]}$ \\
        \textsc{Unit Conversion} & $\mathsf{[M]}$ and $\mathsf{[M]}$ are $\mathsf{[MASK]}$ value \\ 
        \textsc{Reference Range Detection} & $\mathsf{[M]}$ of $\mathsf{[ENT]}$ is $\mathsf{[MASK]}$ \\
        \hline 
        
    \end{tabular}
    }
    \caption{Templates which we used for data generation. $\mathsf{[M]}$,$\mathsf{[LoM]}$, and $\mathsf{[ENT]}$ are the placeholder for the measurement, the list of measurements, and the biomedical entity, respectively.}
    \label{tab:template}
\end{table}

%% file: 03Experiments.tex
\input{Table/02probing}
\par \noindent \textbf{Probing Setup}
\label{subsec:exp_setup}
We formulated MSTs as a Cloze test~\cite{talmor-etal-2020-olmpics} to fully utilize the knowledge captured by masked language modeling (MLM).
Specifically, a PLM received the masked inputs given in Table~\ref{tab:example}, and the MLM head output the probability distribution of the answer candidates for \texttt{[MASK]}.
Among the answer candidates, we chose the one with the highest probability as the final prediction.

We probed four transformer-based PLMs. BERT~\cite{devlin-etal-2019-bert} and ALBERT~\cite{Lan2020ALBERT} were trained on Wikipedia articles and Book Corpus. BioBERT~\cite{lee2020biobert} was trained on biomedical articles from PubMed abstracts, and BlueBERT~\cite{peng-etal-2020-empirical} used both clinical (MIMIC-III~\cite{johnson2016mimic}) and biomedical (PubMed abstracts) corpus for pre-training.
We also tested a randomly initialized transformer encoder (i.e. Scratch) to evaluate the difficulty of our MSTs.
For each model, we did not update the parameters during training, except for the MLM head in the last transformer layer.
In all tasks, the models were trained with three random seeds and we report the mean classification accuracy for all the probing tasks.
Appendix~\ref{sec:appA_MAT_details} provides further details on training and evaluation.

\par \noindent \textbf{Data Preparation}
\label{subsec:data_prep}
We manually crafted templates in Table~\ref{tab:template} that contained at most two slots for measurements and \texttt{[MASK]} token for an answer.
We instantiated \texttt{[M]} and \texttt{[LoM]} by sampling the measurement and the list of measurements, respectively.
For measurement sampling, we independently sampled a number and a unit and then combined them.
Specifically, we sampled units from the predefined set in Table~\ref{tab:list_unit} which consists of SI units and some units in MIMIC-III.

The numbers in the training dataset were sampled from $[10^{-2}, 10^{2})$.
For evaluation, we constructed two evaluation datasets: 1) \textit{Interpolation} sampled numbers from the same range as the training dataset; 2) \textit{Extrapolation} sampled numbers from $[10^{-3}, 10^{3})$.
Note that we did not consider the numbers outside the range $[10^{-3}, 10^{3})$, because many of the unit prefixes are in the power of thousands.
\citet{zhang-etal-2020-language-embeddings} reported that representing numbers in scientific notation made it easier for the language model to capture the scale of numbers.
Following this observation, we tested two different number notations: \textit{decimal} and \textit{scientific}. For example, 32.6 can be represented as 32.6 and 3.26E+01 in decimal and scientific notation, respectively.
We randomly varied the number of digits after the decimal point between zero and three, and the significant digits were maintained after converting the number notation.

For \textsc{reference range detection}, we collected biomedical entities from six tables in MIMIC-III (INPUT, OUTPUT, LAB, PRESCRIPTION, PROCEDURE, and CHART) and chose the subset.

We report the number of samples and the distribution of labels for each MST in Table~\ref{tab:stat}.


%% file: Table/02probing.tex
\begin{table*}[t]
    \centering
    \resizebox{2.0\columnwidth}{!}{%
    \begin{tabular}{lccccccccccc}
    \hline
\multicolumn{2}{c}{\multirow{2}{*}{Task}} & \multicolumn{6}{c}{\textsc{Measurement Comparison}} & \multicolumn{2}{c}{\multirow{2}{*}{\textsc{Unit}}} & \multicolumn{2}{c}{\multirow{2}{*}{\textsc{Ref}}} \\
\cline{3-8}
& & \multicolumn{2}{c}{\textsc{Comp}} & \multicolumn{2}{c}{\textsc{Arg}} & \multicolumn{2}{c}{\textsc{Sort}} & & & & \\
\hline
Model & Notation & $in$ & $ex$ & $in$ & $ex$   & $in$ & $ex$ & $in$ & $ex$  & $in$ & $ex$\\
\hline
\multirow{2}{*}{ALBERT}  
& \textit{Sci}  & 81.2 & 77.3 & \textbf{60.4} & \textbf{58.0} & 78.2 & \textbf{76.5} & 48.6 & 49.9 & 71.9 & 59.9 \\
& \textit{Deci} & 81.8 & 72.1 & 57.1 & 50.5 & \textbf{82.5} & 74.3 & 61.5 & 56.2 & 71.1 & 61.0 \\
\hline
\multirow{2}{*}{BERT}     
& \textit{Sci}  & 73.3 & 72.4 & 55.1 & 52.2 & 45.6 & 45.0 & 52.7 & 51.2 & 73.5 & 64.3 \\
& \textit{Deci} & 81.4 & 77.0 & \textbf{60.9} & 54.3 & 54.9 & 54.5 & 61.9 & 59.2 & 77.2 & 67.5 \\
\hline
\multirow{2}{*}{BioBERT}  
& \textit{Sci}  & 82.7 & 82.3 & 55.0 & 54.4 & 68.2 & 69.1 & 58.7 & 57.3 & 81.3 & 63.7 \\
& \textit{Deci} & \textbf{90.1} & \textbf{88.0} & 59.0 & \textbf{57.6} & 77.3 & 73.0 & \textbf{73.0} & \textbf{70.5} & \textbf{87.0} & 64.2 \\
\hline
\multirow{2}{*}{BlueBERT} 
& \textit{Sci}  & 77.3 & 76.3 & 46.9 & 46.9 & 63.6 & 64.3 & 53.0 & 51.3 & 73.6 & 65.4 \\
& \textit{Deci} & 74.6 & 73.2 & 57.0 & 55.5 & 73.0 & 68.0 & 59.2 & 57.1 & 77.1 & \textbf{69.0} \\
\hline
\multirow{2}{*}{Scratch} 
& \textit{Sci}  & 50.9& 50.8& 40.2& 37.1& 33.3& 33.8& 52.5& 50.7& 66.3& 60.8\\
& \textit{Deci} & 57.7& 51.3& 44.3& 43.0& 33.3& 33.7& 56.8& 53.9& 62.6& 65.0\\
\hline
\end{tabular}
}
\caption{Test-set results on MSTs. We report the classification accuracy on \textit{interpolation} ($in$) and \textit{extrapolation} ($ex$) test dataset. \textsc{comp}, \textsc{arg}, \textsc{sort}, \textsc{unit}, and \textsc{ref} are abbreviations of \textsc{comparison}, \textsc{argmin/max}, \textsc{sorting, }\textsc{Unit Conversion}, and \textsc{Reference Range Detection}, respectively. \textit{Sci} and \textit{Deci} stand for scientific and decimal notations, respectively.}
\label{tab:probing}
\end{table*}

%% file: 04Results.tex
\input{Table/03paraphrase}
\input{Table/04rule_conversion}
\label{subsec:measure_test}
\par \noindent \textbf{Measuring Skills of PLMs}
Table~\ref{tab:probing} shows the results of MSTs stated in Section~\ref{sec:Probing}.

PLMs performed reasonably well on \textsc{comparison}, \textsc{sorting}, and \textsc{reference range detection}, but struggled considerably on \textsc{argmin/max} and \textsc{unit conversion} tasks.
This shows that some measuring skills are difficult to learn from an LM objective.
Similar to previous NRoT studies~\cite{wallace-etal-2019-nlp, pal-baral-2021-investigating-numeracy}, PLMs often failed to successfully extrapolate to values outside the training range.
Further, in most cases, MST results got worse when we represented numbers in scientific notation.

We observed that BioBERT outperformed other PLMs in \textsc{unit conversion}, \textsc{reference range detection}, and \textsc{comparison}, and showed comparable performance in the rest of the MSTs.
Compared to BioBERT, BlueBERT was pre-trained on a larger volume of biomedical text, but showed worse performance.
This shows that pre-training on measurement-rich corpora assists the model in acquiring measuring skills, but further training on the noisy clinical text could harm it when performing reasoning over measurements.
We also found that ALBERT outperformed its competitors in \textsc{Sorting} even though it performed the same or worse on other tasks.
This may be because ALBERT benefits from its sentence order prediction (SOP) objective, which predicts the ordering of two consecutive segments of text.


\par \noindent \textbf{Effect of using Different Prompts}
One can expect that the choice of prompt has an impact on the results, and recent studies~\cite{10.1162/tacl_a_00324, petroni-etal-2019-language} support this. 
To see whether the results in Table~\ref{tab:probing} are maintained as the prompt differs, we trained and evaluated PLMs on three distinct sets of prompts: \textsc{context}, \textsc{UoM}, and \textsc{label}.
Specifically, \textsc{context}, \textsc{UoM}, and \textsc{label} examine how consistent MST results are against various linguistic expressions of prompts, the set of unique UoMs in the dataset, and the choice of answer candidates, respectively. 
Note that we considered answer candidates as part of the prompt, since the prompt determines the set of correct answers.

For \textsc{context}, we manually created four additional templates that have the same meaning as the original template in Table~\ref{tab:template}.
For \textsc{uom}, we used only a subset of units g, l, m, and s, which appear frequently in the general text. 
For \textsc{label}, we included synonyms of the label as answer candidates. 
For example, "\textit{less}", "\textit{smaller}", and "\textit{lower}" are the answers for the prompt "1.59mg is \texttt{[MASK]} than 3.8mg.".
More details of the experiments are in the Appendix~\ref{sec:appB_paraphrase}.


The results with the decimal notation are shown in Table~\ref{tab:paraphrase}. 
We can see that the results vary with the choice of prompt, indicating that PLMs are indeed sensitive to it.
However, we found that MST performance maintains a similar tendency in every experiment: BioBERT works well on \textsc{comparsion}, \textsc{unit conversion}, and \textsc{reference range detection}, and ALBERT works well on \textsc{sorting}.


\input{Figure/scale_embedding}
\input{Table/05scale_emb}
\par \noindent \textbf{Rule-based Conversion of Measurements}
Measurements exhibit a certain pattern, regardless of the domain, because of a global standard: the International System of Units (SI).
Thus, we can manually detect and convert all units in the text without difficulty.
Then, it is natural to wonder if converting all units based on rules is easier than making the language model understand the system of measurement.
To answer this question, we tested the rule-based conversion that detects measurements with the regular expression and converts them into a prefix-free form. 
For example, the sentence "2.5mg is \texttt{[MASK]} than 3.8g" is converted to "0.0025g is \texttt{[MASK]} than 3.8g" after the rule-based conversion.
We examined the rule-based conversion on \textsc{measurement comparison} and \textsc{reference range detection}.

The results with the decimal notation are shown in Table~\ref{tab:manual_conversion}.
The rule-based conversion increased \textsc{measurement comparison} performance because the converted \textsc{measurement comparison} does not require an understanding of unit conversion to solve the problem.
However, it can be seen that almost all models became worse on \textsc{reference range detection}. 
This shows that the knowledge about the reference range is highly correlated with the specific UoM. 
Thus, the rule-based conversion is a suboptimal choice if we want to utilize the domain knowledge embedded in PLMs.


\par \noindent \textbf{Scale Embedding and its Effect}
In Section~\ref{subsec:measure_test}, we observed that none of the PLMs showed a perfect understanding of each MST.
We suspect that such a gap originates in the deficiency of PLM's ability to extract numerical values from measurements and compare their magnitudes.
To this end, we propose \textit{scale embedding}, an additional embedding that provides the model with the information of the position and scale of numbers in the input text.

As described in Figure~\ref{fig:scaleEmb}, we incrementally assigned the index to each token from the end to the beginning of a sentence.
If we encounter a token that is not included in the numerical value, then we reset the index to zero and keep assigning the index zero to tokens until another numerical value appears.
We distinguished between numerical and nonnumerical subwords using the regular expression. 
Note that we trained only the scale embedding and MLM head while freezing other pre-trained weights of the language model.
This allows us to adapt the model to any numerical reasoning tasks simply by plugging a different scale embedding into them.

Table~\ref{tab:scaleEmb} shows the MST results after the scale embedding is applied to all models, where we can see significantly improved test results, even for \textsc{Argmin/max} and \textsc{unit conversion}.\footnote{The full set of experimental results are shown in Table~\ref{tab:scaleEmb_full} in the Appendix.}
Note that the scale embedding is minimally effective for Scratch, except for \textsc{comparison}.
This shows that solving our MSTs requires more than just simple embeddings, and a PLM that understands context is an essential element.

%% file: Table/03paraphrase.tex
\begin{table*}[t]
    \centering
    \resizebox{2.0\columnwidth}{!}{%
    \begin{tabular}{clcccccccccc}
    \hline
\multicolumn{2}{c}{\multirow{2}{*}{Task}} & \multicolumn{6}{c}{\textsc{Measurement Comparison}} & \multicolumn{2}{c}{\multirow{2}{*}{\textsc{Unit}}} & \multicolumn{2}{c}{\multirow{2}{*}{\textsc{Ref}}} \\
\cline{3-8}
& & \multicolumn{2}{c}{\textsc{Comp}} & \multicolumn{2}{c}{\textsc{Arg}} & \multicolumn{2}{c}{\textsc{Sort}} & & & & \\
\hline
Prompt Set & \multicolumn{1}{c}{Model} & $in (\Delta)$ & $ex (\Delta)$ & $in (\Delta)$ & $ex (\Delta)$   & $in (\Delta)$ & $ex (\Delta)$ & $in (\Delta)$ & $ex (\Delta)$  & $in (\Delta)$ & $ex (\Delta)$\\
\hline
\multirow{4}{*}{\textsc{Label}}   
& ALBERT   & 78.7 (-3.1) & 70.8 (-1.3) & 40.9 (-16.2)   & 33.1 (-17.4)  & 73.6 (-8.9)   & 67.2 (-7.1)   & 55.5 (-6.0)   & 56.0 (-0.2)   & 51.1 (-20.0)  & 36.0 (-25.0) \\
& BERT     & 73.1 (-8.3) & 70.8 (-6.2) & 54.0 (-6.9)    & 50.7 (-3.6)   & 54.0 (-0.9)   & 54.3 (-0.2)   & 56.6 (-5.3)   & 55.0 (-4.2)   & 40.3 (-36.9)  & 13.7 (-53.8) \\
& BioBERT  & 82.8 (-7.3) & 80.2 (-7.8) & 56.7 (-2.3)    & 55.7 (-1.9)   & 66.4 (-10.9)  & 62.6 (-10.4)  & 61.9 (-11.1)  & 60.4 (-10.1)  & 69.1 (-17.9)  & 59.6 (-4.6) \\
& BlueBERT & 75.0 (0.4) & 69.7 (-3.5) & 56.9 (-0.1)    & 55.3 (-0.2)   & 70.1 (-2.9)   & 66.6 (-1.4)   & 56.4 (-2.8)   & 54.9 (-2.2)   & 76.4 (-0.7)   & 70.6 (1.6) \\
\hline
\multirow{4}{*}{\textsc{Context}} 
& ALBERT   & 67.8 (-14.0) & 61.8 (-10.3)  & 49.1 (-8.0) & 43.5 (-7.0) & 72.3 (-10.2)  & 68.2 (-6.1)   & 50.4 (-11.1)  & 50.5 (-5.7)   & 65.8 (-5.3)   & 56.9 (-4.1) \\
& BERT     & 70.2 (-11.2) & 67.9 (-9.1)   & 52.4 (-8.5) & 47.1 (-7.2) & 51.8 (-3.1)   & 50.6 (-3.9)   & 56.1 (-5.8)   & 55.2 (-4.0)   & 66.4 (-10.8)  & 63.2 (-4.3) \\
& BioBERT  & 80.7 (-9.4)  & 78.4 (-9.6)   & 58.1 (-0.9) & 55.9 (-1.7) & 73.6 (-3.7)   & 69.7 (-3.3)   & 60.4 (-12.6)  & 59.3 (-11.2)  & 75.2 (-11.8)  & 64.8 (0.6) \\
& BlueBERT & 71.5 (-3.1)  & 65.6 (-7.6)   & 51.9 (-5.1) & 48.5 (-7.0) & 60.6 (-12.4)  & 56.4 (-11.6)  & 50.7 (-8.5)   & 50.7 (-6.4)   & 67.4 (-9.7)   & 64.9 (-4.1) \\
\hline
\multirow{4}{*}{\textsc{UoM}}     
& ALBERT   & 87.9 (6.1)   & 80.2 (8.1)    & 71.0 (13.9)   & 58.8 (8.3)    & 90.5 (8.0)    & 85.5 (11.2)   & 64.2 (2.7) & 56.9 (0.7) & N/A    & N/A    \\
& BERT     & 90.0 (8.6)   & 87.2 (10.2)   & 69.4 (8.5)    & 68.2 (13.9)   & 67.1 (12.2)   & 63.0 (8.5)    & 67.2 (5.3) & 64.8 (5.6) & N/A    & N/A    \\
& BioBERT  & 96.5 (6.4)   & 94.8 (6.8)    & 72.1 (13.1)   & 69.3 (11.7)   & 84.0 (6.7)    & 79.6 (6.6)    & 82.5 (9.5) & 77.9 (7.4) & N/A    & N/A    \\
& BlueBERT & 88.3 (13.7)  & 83.8 (10.6)   & 66.5 (9.5)    & 63.8 (8.3)    & 76.5 (3.5)    & 71.7 (3.7)    & 66.6 (7.4) & 62.4 (5.3) & N/A    & N/A   \\
\hline
\end{tabular}
}
\caption{Test-set results on different sets of prompts. We report the classification accuracy and the performance difference ($\Delta$). We obtain $\Delta$ by subtracting the results in Table~\ref{tab:probing} from this table.}
\label{tab:paraphrase}
\end{table*}

%% file: Table/04rule_conversion.tex
\begin{table*}[t]
    \centering
    \resizebox{2.0\columnwidth}{!}{%
    \begin{tabular}{lccccccccc}
    \hline
\multicolumn{1}{c}{\multirow{2}{*}{Task}} & \multicolumn{6}{c}{\textsc{Measurement Comparison}} & \multicolumn{2}{c}{\multirow{2}{*}{\textsc{Ref}}}\\
\cline{2-7}
& \multicolumn{2}{c}{\textsc{Comp}} & \multicolumn{2}{c}{\textsc{Arg}} & \multicolumn{2}{c}{\textsc{Sort}} & & \\
\hline
Model & $in (\Delta)$ & $ex (\Delta)$ & $in (\Delta)$ & $ex (\Delta)$   & $in (\Delta)$ & $ex (\Delta)$ & $in (\Delta)$ & $ex (\Delta)$\\
\hline & \\[-2.5ex]
ALBERT   & 87.5 (5.7) & 85.3 (13.2) & 71.8 (14.7) & 68.2 (17.7) & 87.3 (4.8) & 85.3 (11.0) & 65.6 (-5.5) & 53.8 (-7.2) \\
BERT     & 88.3 (6.9) & 86.8 (9.8) & 59.3 (-1.6) & 60.6 (6.3) & 77.5 (22.6) & 77.5 (23.0) & 69.6 (-7.6) & 64.2 (-3.3) \\
BioBERT  & 94.7 (4.6) & 93.5 (5.5) & 70.7 (11.7) & 68.3 (10.7) & 83.6 (6.3) & 82.9 (9.9) & 77.6 (-9.4) & 65.7 (1.5) \\
BlueBERT & 88.7 (14.1) & 86.1 (12.9) & 64.8 (7.8) & 60.9 (5.4) & 75.2 (2.2) & 74.2 (6.2) & 70.3 (-6.8) & 66.0 (-3.0) \\
\hline
\end{tabular}
}
\caption{Test-set results on rule-based conversion experiments. We report the classification accuracy and the performance difference ($\Delta$).}
\label{tab:manual_conversion}
\end{table*}

%% file: Figure/scale_embedding.tex
\begin{figure*}[h]
    \centering
    \includegraphics[width=2.0\columnwidth]{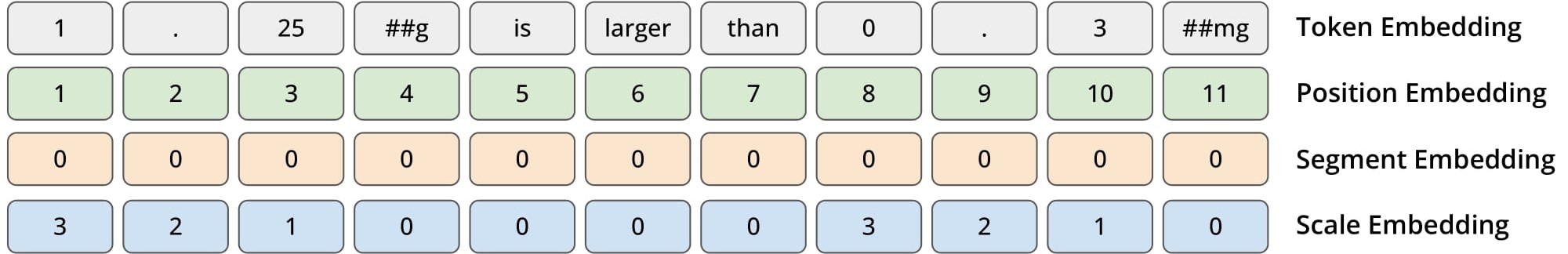}
    \caption{Our scale embedding. }
    \label{fig:scaleEmb}
\end{figure*}

%% file: Table/05scale_emb.tex
\begin{table*}[t]
    \centering
    \resizebox{2.0\columnwidth}{!}{%
    \begin{tabular}{lccccccccccc}
    \hline
\multicolumn{1}{c}{\multirow{2}{*}{Task}} & \multicolumn{6}{c}{\textsc{Measurement Comparison}} & \multicolumn{2}{c}{\multirow{2}{*}{\textsc{Unit}}} & \multicolumn{2}{c}{\multirow{2}{*}{\textsc{Ref}}} \\
\cline{2-7}
& \multicolumn{2}{c}{\textsc{Comp}} & \multicolumn{2}{c}{\textsc{Arg}} & \multicolumn{2}{c}{\textsc{Sort}} & & \\
\hline
Model & $in (\Delta)$ & $ex (\Delta)$ & $in (\Delta)$ & $ex (\Delta)$   & $in (\Delta)$ & $ex (\Delta)$ & $in (\Delta)$ & $ex (\Delta)$  & $in (\Delta)$ & $ex (\Delta)$\\
\hline & \\[-2.5ex]
ALBERT   & 92.9 (11.1) & 78.9 (6.8) & 73.6 (16.5) & 63.5 (13.0) & 92.9 (10.4) & 86.8 (12.5) & 75.7 (14.2)  & 68.0 (11.8) & 83.5 (12.4) & 63.9 (2.9) \\
BERT     & 95.9 (14.5) & 89.0 (12.0) & 79.8 (18.9) & 75.6 (21.3) & 91.7 (36.8) & 90.9 (36.4) & 87.9 (26.0) & 80.2 (21.0) & 95.3 (18.1) & 62.6 (-4.9) \\
BioBERT  & 98.4 (8.3) & 93.2 (5.2) & 85.9 (26.9) & 83.0 (25.4) & 94.0 (16.7) & 93.0 (20.0) & 90.1 (17.1) & 85.7 (15.2) & 98.4 (11.4) & 61.9 (-2.3) \\
BlueBERT & 97.4 (22.8) & 88.1 (14.9) & 75.9 (18.9) & 67.7 (12.2) & 91.8 (18.8) & 90.3 (22.3) & 80.0 (20.8) & 76.2 (19.1) & 94.3 (17.2) & 66.1 (-2.9) \\
Scratch  & 70.2 (12.5) & 60.3 (9.0) & 45.5 (1.2) & 44.1 (1.1) & 33.2 (-0.1) & 33.8 (0.1) & 60.3 (3.5) & 56.1 (2.2) & 69.0 (6.4) & 66.3 (1.3) \\
\hline
\end{tabular}
}
\caption{Effect of scale embedding on MSTs. We report the classification accuracy and performance improvement ($\Delta$) after applying scale embedding.}
\label{tab:scaleEmb}
\end{table*}

%% file: 08RelatedWorks.tex
Over the years, numerical reasoning has been an active research area. 
Some works investigate the numeracy of static word embeddings~\citep{naik-etal-2019-exploring}, contextualized language embeddings~\citep{wallace-etal-2019-nlp}, and multilingual words~\citep{johnson-etal-2020-probing}.
\citet{wallace-etal-2019-nlp} shows that ELMo, BERT, and GloVe embeddings are capable of capturing numeracy, but only within the range of numbers seen during training. 
GenBERT~\citep{geva-etal-2020-injecting}, NumGPT~\citep{jin2021numgpt}, and NT5~\citep{yang2021nt5} focus on incorporating arithmetic skills into pre-trained models. 
Another task that deals with numerical quantities is measurement estimation.
VerbPhysics~\citep{forbes-choi-2017-verb} proposes the dataset to compare the relative scales between the physical attributes of various
objects. 
DoQ~\citep{elazar-etal-2019-large} provides an empirical distribution over possible values of quantitative attributes. 
\citet{zhang-etal-2020-language-embeddings} tests that NLP models contain information about the scalar magnitudes of physical objects.
Although previous studies probed numerical reasoning over numeral and physical attributes, no attempt has been made to investigate reasoning over measurements.

%% file: 05Conclusion.tex
To the best of our knowledge, our study is the first to investigate reasoning over measurements.
Our analysis shows that PLMs lack the capability required for reasoning over measurements.
We proposed a scale embedding approach that provides information on the position and scale of numbers, and it significantly increases the MST performance. 

%% file: 07Limitation.tex
Our scale embedding can make mistakes when the unit itself contains numbers (e.g. mg/100ml). Therefore, scale embedding should not be applied to UoM containing numbers through exception handling.

Our work will be largely affected by the created prompts. 
If the prompt is not obvious for PLMs to understand, although they have such reasoning ability, they may not give the correct answer.
To mitigate this problem, we conducted experiments with different sets of prompts in Section~\ref{sec:Results} and showed that the results maintain their tendency across the prompts.
Despite these efforts, it is still unclear what the optimal choice of the prompt is. 
We remain this problem as a future work.

%% file: 06Appendix.tex

\section{Further Details of Measuring Skill Test}
\label{sec:appA_MAT_details}
\subsection{Data Statistics}
Table~\ref{tab:stat} shows the statistics of MSTs we used for experiments.

\subsection{Training and Evaluation}
The BERT configuration of all models is the same as the base model (L=12, H=768, A=12, Total Parameters=110M) in~\cite{devlin-etal-2019-bert}. Maximum sequence length is 512.

We trained the model with batch size 256 for 30 epochs. We used the Adam optimizer for training. The learning rate started from 5e-5 and linearly decayed towards 1e-8.
We early stopped the training when the validation accuracy did not increase for 2 epochs. The batch size for evaluation is 128, and other settings are the same as training. We found the optimal hyperparameters using the grid search, where we evaluated the learning rate [1e-5, 2e-5, 5e-5, 1e-4], batch size [16,32,64,128].

\input{AppTable/01list_of_units}
\input{AppTable/02data_statistics}

\section{More Details of Prompt Sets}
\label{sec:appB_paraphrase}
The results with both decimal and scientific notation are shown in Table~\ref{tab:paraphrase_full}.

\subsection{\textsc{label}}
Inspired by~\citet{yuan-neurips-2021-bartscore}, we included synonyms as an answer to make the prompt diverse.
We used the website \url{https://www.wordhippo.com/} to search for synonyms.
Among the search results, we chose two words that match the context.
We report the list of synonyms in Table~\ref{tab:para_label}.

\subsection{\textsc{context}}
If the context differs from what PLM saw during pre-training, then PLMs will struggle to solve MSTs even if they understand the measuring skills.
To mitigate this, we prepared four additional prompts with the same meaning.
Additional prompts are listed in Table~\ref{tab:para_template}.

\subsection{\textsc{uom}}
In the general domain, some UoMs listed in Table~\ref{tab:list_unit} rarely appear in the context. For example, international units per liter (IU/l) is frequently used in pharmacology, but not in other scientific articles.
Therefore, we can wonder if some rare biomedical units disrupt the understanding of general domain PLMs (e.g., BERT and ALBERT).
To answer this question, we replaced all UoMs in the dataset with the commonly used UoMs: g, l, m, and s.

\section{Additional Results on Rule-based Conversion}
\label{sec:appC_rule}
Table~\ref{tab:manual_conversion_full} describes the complete set of MST results after applying rule-based conversion.

\section{Additional Results on Scale Embedding}
\label{sec:appD_scale}
Table~\ref{tab:scaleEmb_full} describes the MST results of scale embedding with decimal and scientific notation.

\section{Experimental Environment}
We trained the models with Google TPU v2-8 and v3-8. 
We used PyTorch 1.10.0~\cite{NEURIPS2019_9015} and Huggingface Transformers~\cite{wolf-etal-2020-transformers} 4.3.3 for experiments. 

\input{AppTable/05paraphrase}
\input{AppTable/03sample_label}

\input{AppTable/04sample_context}
\input{AppTable/06rule_conversion}
\input{AppTable/07scale_emb}

%% file: AppTable/01list_of_units.tex
\begin{table}[t]
    \centering
    \resizebox{1.0\columnwidth}{!}{%
    \begin{tabular}{l|l}
        \hline 
        \multicolumn{2}{c}{\textbf{UoM}}\\
        \hline
        \textbf{w/o Prefix} & \multicolumn{1}{c}{\textbf{w/ Prefix}}\\
        \hline 
        m & m, cm, mm, µm, nm \\
        A & A, mA, µA, nA \\
        K & K, mK, µK \\
        M & M, mM, µM, nM \\
        Eq/l & Eq/l, mEq/l, µEq/l, mEq/ml, mEq/µl\\
        g/l & g/l, mg/l, µg/l, mg/dl, g/dl, µg/dl, ng/dl, g/ml, mg/ml \\
        IU/l & IU/l, IU/ml, mIU/ml, µIU/ml, mIU/l, µIU/l, IU/µl, mIU/µl \\
        U/l & U/l, U/ml, U/µl \\
        l/min & l/min, dl/min, ml/min, µl/min \\
        \#/l & \#/dl, \#/ml, \#/µl \\
        k/l & k/dl, k/ml, k/µl \\
        l & l, dl, ml, µl, nl, pl, fl \\
        g & g, mg, µg, ng, pg, fg \\
        s & s, ms, µs, ns \\
        m/hr & m/hr, cm/hr, mm/hr, µm/hr \\
        l/hr & l/hr, dl/hr, ml/hr, µl/hr \\ 
        \hline 
    \end{tabular}
    }
\caption{List of units used for data generation.}
\label{tab:list_unit}
\end{table}

%% file: AppTable/02data_statistics.tex
\begin{table*}[t]
    \centering
    \resizebox{2.0\columnwidth}{!}{%
    \begin{tabular}{l|c|c|l|l}
        \hline 
        \textbf{Task} & \textbf{Split} & \textbf{Number Range} & \textbf{\# Samples} & \textbf{Label Distribution}\\
        \hline 
\multirow{5}{*}{\textsc{comparison}} & train                                     & \textit{interpolation}& 299,394 & smaller: 0.5, larger: 0.5                           \\
                                                      \cline{2-5}
                                                      & \multirow{2}{*}{valid} & \textit{interpolation}& 29,986  & smaller: 0.498, larger: 0.502                       \\
                                                      &                        & \textit{extrapolation} & 30,000  & smaller: 0.495, larger: 0.505                       \\
                                                      \cline{2-5}
                                                      
                                                      & \multirow{2}{*}{test}  & \textit{interpolation}& 29,988  & smaller: 0.501, larger: 0.499                       \\
                                                      &                        & \textit{extrapolation} & 30,000  & smaller: 0.501, larger: 0.499                       \\
\hline
\multirow{5}{*}{\textsc{argmin/max}} & train                                    & \textit{interpolation}& 300,000 & smallest:0.333, middle: 0.334, largest: 0.333       \\
\cline{2-5}
                                                      & \multirow{2}{*}{valid} & \textit{interpolation}& 30,000  & smallest:0.333, middle: 0.334, largest: 0.333       \\
                                                      &                        & \textit{extrapolation} & 30,000  & smallest:0.333, middle: 0.334, largest: 0.333       \\
                                                      \cline{2-5}
                                                      & \multirow{2}{*}{test}  & \textit{interpolation}& 30,000  & smallest:0.332, middle: 0.335, largest: 0.333       \\
                                                      &                        & \textit{extrapolation} & 30,000  & smallest:0.335, middle: 0.332, largest: 0.333       \\
\hline
\multirow{5}{*}{\textsc{sorting}} & train                                       & \textit{interpolation}& 300,000 & decreasing: 0.333, random: 0.332, increasing: 0.335 \\
                                                      \cline{2-5}
                                                      & \multirow{2}{*}{valid} & \textit{extrapolation} & 30,000  & decreasing: 0.333, random: 0.332, increasing: 0.335 \\
                                                      &                        & \textit{extrapolation} & 30,000  & decreasing: 0.337, random: 0.332, increasing: 0.331 \\
                                                      \cline{2-5}
                                                      & \multirow{2}{*}{test}  & \textit{interpolation}& 30,000  & decreasing: 0.337, random: 0.332, increasing: 0.331 \\
                                                      &                        & \textit{extrapolation} & 30,000  & decreasing: 0.328, random: 0.339, increasing: 0.333 \\
\hline
\multirow{5}{*}{\textsc{unit conversion}} & train                              & \textit{interpolation}& 259,588 & same: 0.489, different: 0.511                       \\
                                                      \cline{2-5}
                                                      & \multirow{2}{*}{valid} & \textit{interpolation}& 23,931  & same: 0.489, different: 0.511                       \\
                                                      &                        & \textit{extrapolation} & 28,814  & same: 0.5, different: 0.5                           \\
                                                      \cline{2-5}
                                                      & \multirow{2}{*}{test}  & \textit{interpolation}& 23,538  & same: 0.483, different: 0.517                       \\
                                                      &                        & \textit{extrapolation} & 28,696  & same: 0.498, different: 0.502                       \\
\hline
\multirow{5}{*}{\textsc{reference range detection}} & train                    & \textit{interpolation}& 201,061 & normal: 0.575, abnormal: 0.425                      \\
                                                      \cline{2-5}
                                                      & \multirow{2}{*}{valid} & \textit{interpolation}& 17,111  & normal: 0.593, abnormal: 0.407                      \\
                                                      &                        & \textit{extrapolation} & 21,212  & normal: 0.618, abnormal: 0.382                      \\
                                                      \cline{2-5}
                                                      & \multirow{2}{*}{test}  & \textit{interpolation}& 16,948  & normal: 0.586, abnormal: 0.414                      \\
                                                      &                        & \textit{extrapolation} & 18,429  & normal: 0.659, abnormal: 0.341 
                                                      \\
    \hline 
    \end{tabular}
    }
\caption{Statistics of MSTs used for experiments.}
\label{tab:stat}
\end{table*}

%% file: AppTable/05paraphrase.tex
\begin{table*}[t]
    \centering
    \resizebox{2.0\columnwidth}{!}{%
    \begin{tabular}{clccccccccccc}
    \hline
\multicolumn{3}{c}{\multirow{2}{*}{Task}} & \multicolumn{6}{c}{\textsc{Measurement Comparison}} & \multicolumn{2}{c}{\multirow{2}{*}{\textsc{Unit}}} & \multicolumn{2}{c}{\multirow{2}{*}{\textsc{Ref}}} \\
\cline{4-9}
& & & \multicolumn{2}{c}{\textsc{Comp}} & \multicolumn{2}{c}{\textsc{Arg}} & \multicolumn{2}{c}{\textsc{Sort}} & & & & \\
\hline
Prompt Set & \multicolumn{1}{c}{Model} & \multicolumn{1}{c}{Notation} & $in (\Delta)$ & $ex (\Delta)$ & $in (\Delta)$ & $ex (\Delta)$   & $in (\Delta)$ & $ex (\Delta)$ & $in (\Delta)$ & $ex (\Delta)$  & $in (\Delta)$ & $ex (\Delta)$\\
\hline & \\[-2.5ex]
\multirow{8}{*}{\textsc{Label}}
    & \multirow{2}{*}{ALBERT}   & \textit{Sci}    & 77.1 (-4.1) & 73.8 (-3.5) & 40.5 (-19.9) & 34.2 (-23.8) & 71.3 (-6.9) & 65.6 (-10.9) & 54.5 (5.9) & 53.0 (3.1) & 51.4 (-20.5) & 34.6 (-25.3) \\
    &                           & \textit{Deci}   & 78.7 (-3.1) & 70.8 (-1.3) & 40.9 (-16.2) & 33.1 (-17.4) & 73.6 (-8.9) & 67.2 (-7.1) & 55.5 (-6.0) & 56.0 (-0.2) & 51.1 (-20.0) & 36.0 (-25.0) \\
\cline{2-13} & \\[-2.5ex]
    & \multirow{2}{*}{BERT}     & \textit{Sci}    & 68.9 (-4.4) & 68.7 (-3.7) & 42.9 (-12.2) & 42.9 (-9.3) & 47.3 (1.7) & 46.4 (1.4) & 53.4 (0.7) & 52.2 (1.0) & 47.4 (-26.1) & 23.0 (-41.3) \\
    &                           & \textit{Deci}   & 73.1 (-8.3) & 70.8 (-6.2) & 54.0 (-6.9) & 50.7 (-3.6) & 54.0 (-0.9) & 54.3 (-0.2) & 56.6 (-5.3) & 55.0 (-4.2) & 40.3 (-36.9) & 13.7 (-53.8) \\
\cline{2-13} & \\[-2.5ex]
    & \multirow{2}{*}{BioBERT}  & \textit{Sci}    & 74.0 (-8.7) & 73.3 (-9.0) & 50.9 (-4.1) & 50.8 (-3.6) & 61.8 (-6.4) & 61.7 (-7.4) & 55.8 (-2.9) & 54.1 (-3.2) & 59.3 (-22.0) & 54.3 (-9.4) \\
    &                           & \textit{Deci}   & 82.8 (-7.3) & 80.2 (-7.8) & 56.7 (-2.3) & 55.7 (-1.9) & 66.4 (-10.9) & 62.6 (-10.4) & 61.9 (-11.1) & 60.4 (-10.1) & 69.1 (-17.9) & 59.6 (-4.6) \\
\cline{2-13} & \\[-2.5ex]
    & \multirow{2}{*}{BlueBERT} & \textit{Sci}    & 77.0 (-0.3) & 76.3 (0.0) & 44.7 (-2.2) & 44.3 (-2.6) & 64.6 (1.0) & 65.1 (0.8) & 52.9 (-0.1) & 51.3 (0.0) & 70.0 (-3.6) & 63.8 (-1.6) \\
    &                           & \textit{Deci}   & 75.0 (0.4) & 69.7 (-3.5) & 56.9 (-0.1) & 55.3 (-0.2) & 70.1 (-2.9) & 66.6 (-1.4) & 56.4 (-2.8) & 54.9 (-2.2) & 76.4 (-0.7) & 70.6 (1.6) \\
\hline & \\[-2.5ex]
\multirow{8}{*}{\textsc{Context}}   
    & \multirow{2}{*}{ALBERT}   & \textit{Sci}    & 68.4 (-12.8) & 65.4 (-11.9) & 49.3 (-11.1) & 48.2 (-9.8) & 71.4 (-6.8) & 69.2 (-7.3) & 49.7 (1.1) & 50.0 (0.1) & 65.7 (-6.2) & 57.2 (-2.7) \\
    &                           & \textit{Deci}   & 67.8 (-14.0) & 61.8 (-10.3)  & 49.1 (-8.0) & 43.5 (-7.0) & 72.3 (-10.2)  & 68.2 (-6.1)   & 50.4 (-11.1)  & 50.5 (-5.7)   & 65.8 (-5.3)   & 56.9 (-4.1) \\
\cline{2-13} & \\[-2.5ex]
    & \multirow{2}{*}{BERT}     & \textit{Sci}    & 65.4 (-7.9) & 64.5 (-7.9) & 46.8 (-8.3) & 44.6 (-7.6) & 44.6 (-1.0) & 43.9 (-1.1) & 52.7 (0.0) & 52.5 (1.3) & 65.4 (-8.1) & 58.9 (-5.4) \\
    &                           & \textit{Deci}   & 70.2 (-11.2) & 67.9 (-9.1)   & 52.4 (-8.5) & 47.1 (-7.2) & 51.8 (-3.1)   & 50.6 (-3.9)   & 56.1 (-5.8)   & 55.2 (-4.0)   & 66.4 (-10.8)  & 63.2 (-4.3) \\
\cline{2-13} & \\[-2.5ex]
    & \multirow{2}{*}{BioBERT}  & \textit{Sci}    & 74.6 (-8.1) & 74.2 (-8.1) & 53.3 (-1.7) & 52.9 (-1.5) & 68.4 (0.2) & 67.9 (-1.2) & 55.0 (-3.7) & 54.7 (-2.6) & 70.8 (-10.5) & 61.2 (-2.5) \\
    &                           & \textit{Deci}   & 80.7 (-9.4)  & 78.4 (-9.6)   & 58.1 (-0.9) & 55.9 (-1.7) & 73.6 (-3.7)   & 69.7 (-3.3)   & 60.4 (-12.6)  & 59.3 (-11.2)  & 75.2 (-11.8)  & 64.8 (0.6) \\
\cline{2-13} & \\[-2.5ex] 
    & \multirow{2}{*}{BlueBERT} & \textit{Sci}    & 68.4 (-8.9) & 67.1 (-9.2) & 40.4 (-6.5) & 40.6 (-6.3) & 56.8 (-6.8) & 56.5 (-7.8) & 50.3 (-2.7) & 50.2 (-1.1) & 66.5 (-7.1) & 60.3 (-5.1) \\
    &                           & \textit{Deci}   & 71.5 (-3.1)  & 65.6 (-7.6)   & 51.9 (-5.1) & 48.5 (-7.0) & 60.6 (-12.4)  & 56.4 (-11.6)  & 50.7 (-8.5)   & 50.7 (-6.4)   & 67.4 (-9.7)   & 64.9 (-4.1) \\
\hline
\multirow{8}{*}{\textsc{UoM}}  & \\[-2.5ex] 
    & \multirow{2}{*}{ALBERT}   & \textit{Sci}    & 92.8 (11.6) & 87.2 (9.9) & 73.0 (12.6) & 68.7 (10.7) & 89.0 (10.8) & 84.6 (8.1) & 60.5 (11.9) & 56.1 (6.2) & N/A & N/A \\
    &                           & \textit{Deci}   & 87.9 (6.1)   & 80.2 (8.1)    & 71.0 (13.9)   & 58.8 (8.3)    & 90.5 (8.0)    & 85.5 (11.2)   & 64.2 (2.7) & 56.9 (0.7) & N/A    & N/A    \\
\cline{2-13} & \\[-2.5ex]
    & \multirow{2}{*}{BERT}     & \textit{Sci}    & 81.5 (8.2) & 78.9 (6.5) & 60.6 (5.5) & 61.8 (9.6) & 50.7 (5.1) & 49.0 (4.0) & 56.5 (3.8) &  53.2 (2.0) &  N/A  &  N/A \\
    &                           & \textit{Deci}   & 90.0 (8.6)   & 87.2 (10.2)   & 69.4 (8.5)    & 68.2 (13.9)   & 67.1 (12.2)   & 63.0 (8.5)    & 67.2 (5.3) & 64.8 (5.6) & N/A    & N/A    \\
\cline{2-13} & \\[-2.5ex]
    & \multirow{2}{*}{BioBERT}  & \textit{Sci}    & 91.3 (8.6) & 90.3 (8.0) & 53.6 (-1.4) & 52.2 (-2.2) & 72.7 (4.5) & 70.5 (1.4) & 66.1 (7.4) & 63.2 (5.9) & N/A & N/A \\
    &                           & \textit{Deci}   & 96.5 (6.4)   & 94.8 (6.8)    & 72.1 (13.1)   & 69.3 (11.7)   & 84.0 (6.7)    & 79.6 (6.6)    & 82.5 (9.5) & 77.9 (7.4) & N/A    & N/A    \\
\cline{2-13} & \\[-2.5ex]
    & \multirow{2}{*}{BlueBERT} & \textit{Sci}    & 85.8 (8.5) & 83.0 (6.7) & 57.9 (11.0) & 57.3 (10.4) & 68.6 (5.0) & 69.4 (5.1) & 57.2 (4.2) & 53.5 (2.2) & N/A & N/A \\
    &                           & \textit{Deci}   & 88.3 (13.7)  & 83.8 (10.6)   & 66.5 (9.5)    & 63.8 (8.3)    & 76.5 (3.5)    & 71.7 (3.7)    & 66.6 (7.4) & 62.4 (5.3) & N/A    & N/A   \\
\hline
\end{tabular}
}
\caption{Test-set results on different sets of prompts. We report the classification accuracy and the performance difference ($\Delta$). We obtain $\Delta$ by subtracting the results in Table~\ref{tab:probing} from this table.}
\label{tab:paraphrase_full}
\end{table*}

%% file: AppTable/03sample_label.tex
\begin{table*}[t]
    \centering
    \resizebox{2.0\columnwidth}{!}{%
    \begin{tabular}{l|l|l}
        \hline 
        \textbf{Task} & \textbf{Answer Candidates} & \textbf{Synonyms}\\
        \hline 
        \multirow{2}{*}{\textsc{comparison}}
         & larger & higher, bigger \\
         & smaller & lower, less \\
        \hline
        \multirow{3}{*}{\textsc{argmin/max}} 
         & largest & biggest, maximum \\
         & middle & medium, intermediate \\
         & smallest & lowest, minimum \\
        \hline
        \multirow{3}{*}{\textsc{sorting}} 
         & increasing & growing, ascending \\
         & random & unclear, confusing \\
         & decreasing & reducing, descending \\
        \hline
        \multirow{2}{*}{\textsc{unit conversion}}
         & same & equal, identical \\
         & different & distinct, unlike \\
        \hline
        \multirow{2}{*}{\textsc{reference range detection}} 
         & normal & regular, safe \\
         & abnormal & irregular, lethal \\
        \hline 
    \end{tabular}
    }
\caption{Synonyms of the answer candidates we used for \textsc{label}.}
\label{tab:para_label}
\end{table*}

%% file: AppTable/04sample_context.tex
\begin{table*}[t]
    \centering
    \resizebox{2.0\columnwidth}{!}{%
    \begin{tabular}{l|l}
        \hline 
        \textbf{Task} & \textbf{Template}\\
        \hline 
        \multirow{5}{*}{\textsc{comparison}}
        & $\mathsf{[M]}$ is $\mathsf{[MASK]}$ than $\mathsf{[M]}$ \\
        & compared to $\mathsf{[M]}$, $\mathsf{[M]}$ is $\mathsf{[MASK]}$ value \\
        & the measurement of control group ($\mathsf{[M]}$) is $\mathsf{[MASK]}$ than $\mathsf{[M]}$ \\
        & comparison: $\mathsf{[M]}$, $\mathsf{[M]}$, result: $\mathsf{[MASK]}$ \\
        & $\mathsf{[M]}$ $\mathsf{[MASK]}$ $\mathsf{[M]}$ \\
        \hline
        \multirow{5}{*}{\textsc{argmin/max}} 
        & The $\mathsf{[MASK]}$ value among $\mathsf{[LoM]}$ is $\mathsf{[M]}$ \\
        & $\mathsf{[M]}$ is the $\mathsf{[MASK]}$ value of $\mathsf{[LoM]}$ \\
        & Among the list of measurements $\mathsf{[LoM]}$, the $\mathsf{[MASK]}$ value is $\mathsf{[M]}$ \\
        & argmin,argmax: $\mathsf{[LoM]}$, $\mathsf{[M]}$, result: $\mathsf{[MASK]}$ \\
        & $\mathsf{[MASK]}$ $\mathsf{[LoM]}$ , $\mathsf{[M]}$ \\
        \hline
        \multirow{5}{*}{\textsc{sorting}} 
        & sort $\mathsf{[LoM]}$ in $\mathsf{[MASK]}$ order is $\mathsf{[LoM]}$ \\
        & arranging $\mathsf{[LoM]}$ in $\mathsf{[MASK]}$ order is $\mathsf{[LoM]}$ \\
        & $\mathsf{[LoM]}$ is obtained by sorting $\mathsf{[LoM]}$ in $\mathsf{[MASK]}$ order \\
        & sort: $\mathsf{[LoM]}$, $\mathsf{[LoM]}$, result: $\mathsf{[MASK]}$ \\
        & $\mathsf{[LoM]}$ $\mathsf{[MASK]}$ $\mathsf{[LoM]}$ \\
        \hline
        \multirow{5}{*}{\textsc{unit conversion}}
        & $\mathsf{[M]}$ and $\mathsf{[M]}$ are the $\mathsf{[MASK]}$ value \\
        & convert $\mathsf{[M]}$ to $\mathsf{[MASK]}$ value, then the result is $\mathsf{[M]}$ \\
        & compare $\mathsf{[M]}$ to $\mathsf{[M]}$, the two are the $\mathsf{[MASK]}$ value \\
        & measurement comparison: $\mathsf{[M]}$, $\mathsf{[M]}$, result: $\mathsf{[MASK]}$ \\
        & $\mathsf{[M]}$ , $\mathsf{[M]}$ $\mathsf{[MASK]}$ \\
        \hline
        \multirow{5}{*}{\textsc{reference range detection}} 
        & $\mathsf{[M]}$ of $\mathsf{[ENT]}$ is $\mathsf{[MASK]}$ \\
        & $\mathsf{[M]}$ of $\mathsf{[ENT]}$ falls into $\mathsf{[MASK]}$ range \\
        & The physician decides $\mathsf{[M]}$ of $\mathsf{[ENT]}$ as $\mathsf{[MASK]}$ \\
        & reference range: $\mathsf{[ENT]}$, $\mathsf{[M]}$, result: $\mathsf{[MASK]}$ \\
        & $\mathsf{[ENT]}$ $\mathsf{[M]}$ $\mathsf{[MASK]}$ \\
        \hline 
    \end{tabular}
    }
\caption{Templates for \textsc{context}. $\mathsf{[M]}$ is the measurement and $\mathsf{[LoM]}$ is the list of measurements.}
\label{tab:para_template}
\end{table*}

%% file: AppTable/06rule_conversion.tex
\begin{table*}[t]
    \centering
    \resizebox{2.0\columnwidth}{!}{%
    \begin{tabular}{lcccccccccc}
    \hline
\multicolumn{2}{c}{\multirow{2}{*}{Task}} & \multicolumn{6}{c}{\textsc{Measurement Comparison}} & \multicolumn{2}{c}{\multirow{2}{*}{\textsc{Ref}}} \\
\cline{3-8}
& & \multicolumn{2}{c}{\textsc{Comp}} & \multicolumn{2}{c}{\textsc{Arg}} & \multicolumn{2}{c}{\textsc{Sort}} & & \\
\hline
Model & Notation & $in (\Delta)$ & $ex (\Delta)$ & $in (\Delta)$ & $ex (\Delta)$   & $in (\Delta)$ & $ex (\Delta)$ & $in (\Delta)$ & $ex (\Delta)$\\
\hline & \\[-2.5ex]
\multirow{2}{*}{ALBERT}   
    &  \textit{Sci}     & 73.8 (-7.4) & 73.0 (-4.3) & 54.7 (-5.7) & 51.0 (-7.0) & 80.4 (2.2) & 77.7 (1.2) & 71.8 (-0.1) & 62.3 (2.4) \\
    &  \textit{Deci}    & 87.5 (5.7) & 85.3 (13.2) & 71.8 (14.7) & 68.2 (17.7) & 87.3 (4.8) & 85.3 (11.0) & 65.6 (-5.5) & 53.8 (-7.2) \\
\hline \\[-2.5ex]
\multirow{2}{*}{BERT}     
    &  \textit{Sci}     & 69.2 (-4.1) & 68.6 (-3.8) & 53.7 (-1.4) & 52.4 (0.2) & 53.4 (7.8) & 53.4 (8.4) & 73.2 (-0.3) & 63.5 (-0.8) \\
    &  \textit{Deci}    & 88.3 (6.9) & 86.8 (9.8) & 59.3 (-1.6) & 60.6 (6.3) & 77.5 (22.6) & 77.5 (23.0) & 69.6 (-7.6) & 64.2 (-3.3) \\
\hline \\[-2.5ex]
\multirow{2}{*}{BioBERT}  
    &  \textit{Sci}     & 80.6 (-2.1) & 80.1 (-2.2) & 50.4 (-4.6) & 47.0 (-7.4) & 69.0 (0.8) & 68.1 (-1.0) & 79.0 (-2.3) & 64.3 (0.6) \\
    &  \textit{Deci}    & 94.7 (4.6) & 93.5 (5.5) & 70.7 (11.7) & 68.3 (10.7) & 83.6 (6.3) & 82.9 (9.9) & 77.6 (-9.4) & 65.7 (1.5) \\
\hline \\[-2.5ex]
\multirow{2}{*}{BlueBERT} 
    &  \textit{Sci}     & 68.3 (-9.0) & 65.8 (-10.5) & 40.2 (-6.7) & 40.1 (-6.8) & 66.9 (3.3) & 67.0 (2.7) & 74.0 (0.4) & 65.0 (-0.4) \\
    &  \textit{Deci}    & 88.7 (14.1) & 86.1 (12.9) & 64.8 (7.8) & 60.9 (5.4) & 75.2 (2.2) & 74.2 (6.2) & 70.3 (-6.8) & 66.0 (-3.0) \\
\hline \\[-2.5ex]
\multirow{2}{*}{Scratch}  
    &  \textit{Sci}     & 58.2 (7.3) & 54.6 (3.8) & 43.0 (2.8) & 41.0 (3.9) & 33.3 (0.0) & 33.7 (-0.1) & 64.7 (-1.6) & 62.8 (2.0) \\
    &  \textit{Deci}    & 78.8 (21.1) & 73.9 (22.6) & 43.5 (-0.8) & 42.9 (-0.1) & 33.2 (-0.1) & 33.9 (0.2) & 63.6 (1.0) & 64.3 (-0.7) \\
\hline
\end{tabular}
}
\caption{Test-set results on rule-based conversion experiments. We report the classification accuracy and the performance difference.}
\label{tab:manual_conversion_full}
\end{table*}

%% file: AppTable/07scale_emb.tex
\begin{table*}[t]
    \centering
    \resizebox{2.0\columnwidth}{!}{%
    \begin{tabular}{lcccccccccccc}
    \hline
\multicolumn{2}{c}{\multirow{2}{*}{Task}} & \multicolumn{6}{c}{\textsc{Measurement Comparison}} & \multicolumn{2}{c}{\multirow{2}{*}{\textsc{Unit}}} & \multicolumn{2}{c}{\multirow{2}{*}{\textsc{Ref}}} \\
\cline{3-8}
& & \multicolumn{2}{c}{\textsc{Comp}} & \multicolumn{2}{c}{\textsc{Arg}} & \multicolumn{2}{c}{\textsc{Sort}} & & \\
\hline
Model & Notation & $in (\Delta)$ & $ex (\Delta)$ & $in (\Delta)$ & $ex (\Delta)$   & $in (\Delta)$ & $ex (\Delta)$ & $in (\Delta)$ & $ex (\Delta)$  & $in (\Delta)$ & $ex (\Delta)$\\
\hline & \\[-2.5ex]
\multirow{2}{*}{ALBERT}  
    &  \textit{Sci}     & 93.4 (12.2) & 86.3 (9.0) & 73.2 (12.8) & 66.0 (8.0) & 92.7 (14.5) & 90.1 (13.6) & 74.8 (26.2) & 61.6 (11.7) & 87.0 (15.1) & 63.4 (3.5) \\
    &  \textit{Deci}    & 92.9 (11.1) & 78.9 (6.8) & 73.6 (16.5) & 63.5 (13.0) & 92.9 (10.4) & 86.8 (12.5) & 75.7 (14.2)  & 68.0 (11.8) & 83.5 (12.4) & 63.9 (2.9) \\
\hline & \\[-2.5ex]
\multirow{2}{*}{BERT}  
    &  \textit{Sci}     & 96.4 (23.1) & 95.0 (22.6) & 80.9 (25.8) & 80.5 (28.3) & 89.8 (44.2) & 89.5 (44.5) & 79.9 (27.2) & 67.5 (16.3) & 92.4 (18.9) & 61.9 (-2.4) \\
    &  \textit{Deci}    & 95.9 (14.5) & 89.0 (12.0) & 79.8 (18.9) & 75.6 (21.3) & 91.7 (36.8) & 90.9 (36.4) & 87.9 (26.0) & 80.2 (21.0) & 95.3 (18.1) & 62.6 (-4.9) \\
\hline & \\[-2.5ex]
\multirow{2}{*}{BioBERT}
    &  \textit{Sci}     & 98.3 (15.6) & 96.3 (14.0) & 81.3 (26.3) & 80.7 (26.3) & 94.0 (25.8) & 93.6 (24.5) & 89.3 (30.6) & 66.7 (9.4) & 96.0 (14.7) & 64.7 (1.0) \\
    &  \textit{Deci}    & 98.4 (8.3) & 93.2 (5.2) & 85.9 (26.9) & 83.0 (25.4) & 94.0 (16.7) & 93.0 (20.0) & 90.1 (17.1) & 85.7 (15.2) & 98.4 (11.4) & 61.9 (-2.3) \\
\hline & \\[-2.5ex]
\multirow{2}{*}{BlueBERT}
    &  \textit{Sci}     & 96.0 (18.7) & 93.1 (16.8) & 76.0 (29.1) & 74.9 (28.0) & 86.2 (22.6) & 85.8 (21.5) & 77.4 (24.4) & 63.3 (12.0) & 91.1 (17.5) & 66.7 (1.3) \\
    &  \textit{Deci}    & 97.4 (22.8) & 88.1 (14.9) & 75.9 (18.9) & 67.7 (12.2) & 91.8 (18.8) & 90.3 (22.3) & 80.0 (20.8) & 76.2 (19.1) & 94.3 (17.2) & 66.1 (-2.9) \\
\hline & \\[-2.5ex]
\multirow{2}{*}{Scratch}  
    &  \textit{Sci}     & 59.5 (8.6) & 57.4 (6.6) & 41.4 (1.2) & 39.9 (2.8) & 33.4 (0.1) & 33.8 (0.0) & 52.5 (0.0) & 50.6 (-0.1) & 80.0 (13.7) & 61.8 (1.0) \\
    &  \textit{Deci}    & 70.2 (12.5) & 60.3 (9.0) & 45.5 (1.2) & 44.1 (1.1) & 33.2 (-0.1) & 33.8 (0.1) & 60.3 (3.5) & 56.1 (2.2) & 69.0 (6.4) & 66.3 (1.3) \\
\hline
\end{tabular}
}
\caption{Effect of scale embedding on MSTs. We report the classification accuracy and performance improvement ($\Delta$) after applying scale embedding.}
\label{tab:scaleEmb_full}
\end{table*}

%% file: emnlp2022.bbl
\begin{thebibliography}{24}
\expandafter\ifx\csname natexlab\endcsname\relax\def\natexlab#1{#1}\fi

\bibitem[{Devlin et~al.(2019)Devlin, Chang, Lee, and
  Toutanova}]{devlin-etal-2019-bert}
Jacob Devlin, Ming-Wei Chang, Kenton Lee, and Kristina Toutanova. 2019.
\newblock \href {https://doi.org/10.18653/v1/N19-1423} {{BERT}: Pre-training of
  deep bidirectional transformers for language understanding}.
\newblock In \emph{Proceedings of the 2019 Conference of the North {A}merican
  Chapter of the Association for Computational Linguistics: Human Language
  Technologies, Volume 1 (Long and Short Papers)}, pages 4171--4186,
  Minneapolis, Minnesota. Association for Computational Linguistics.

\bibitem[{Elazar et~al.(2019)Elazar, Mahabal, Ramachandran, Bedrax-Weiss, and
  Roth}]{elazar-etal-2019-large}
Yanai Elazar, Abhijit Mahabal, Deepak Ramachandran, Tania Bedrax-Weiss, and Dan
  Roth. 2019.
\newblock \href {https://doi.org/10.18653/v1/P19-1388} {How large are lions?
  inducing distributions over quantitative attributes}.
\newblock In \emph{Proceedings of the 57th Annual Meeting of the Association
  for Computational Linguistics}, pages 3973--3983, Florence, Italy.
  Association for Computational Linguistics.

\bibitem[{Forbes and Choi(2017)}]{forbes-choi-2017-verb}
Maxwell Forbes and Yejin Choi. 2017.
\newblock \href {https://doi.org/10.18653/v1/P17-1025} {Verb physics: Relative
  physical knowledge of actions and objects}.
\newblock In \emph{Proceedings of the 55th Annual Meeting of the Association
  for Computational Linguistics (Volume 1: Long Papers)}, pages 266--276,
  Vancouver, Canada. Association for Computational Linguistics.

\bibitem[{Geva et~al.(2020)Geva, Gupta, and Berant}]{geva-etal-2020-injecting}
Mor Geva, Ankit Gupta, and Jonathan Berant. 2020.
\newblock \href {https://doi.org/10.18653/v1/2020.acl-main.89} {Injecting
  numerical reasoning skills into language models}.
\newblock In \emph{Proceedings of the 58th Annual Meeting of the Association
  for Computational Linguistics}, pages 946--958, Online. Association for
  Computational Linguistics.

\bibitem[{Jiang et~al.(2020)Jiang, Xu, Araki, and
  Neubig}]{10.1162/tacl_a_00324}
Zhengbao Jiang, Frank~F. Xu, Jun Araki, and Graham Neubig. 2020.
\newblock \href {https://doi.org/10.1162/tacl_a_00324} {{How Can We Know What
  Language Models Know?}}
\newblock \emph{Transactions of the Association for Computational Linguistics},
  8:423--438.

\bibitem[{Jin et~al.(2021)Jin, Jiang, Wang, Liu, Wang, Ren, and
  Qu}]{jin2021numgpt}
Zhihua Jin, Xin Jiang, Xingbo Wang, Qun Liu, Yong Wang, Xiaozhe Ren, and Huamin
  Qu. 2021.
\newblock \href {http://arxiv.org/abs/2109.03137} {Numgpt: Improving numeracy
  ability of generative pre-trained models}.
\newblock \emph{arXiv preprint arXiv:2109.03137}.

\bibitem[{Johnson et~al.(2016)Johnson, Pollard, Shen, Lehman, Feng, Ghassemi,
  Moody, Szolovits, Celi, and Mark}]{johnson2016mimic}
Alistair E~W Johnson, Tom~J Pollard, Lu~Shen, Li-Wei~H Lehman, Mengling Feng,
  Mohammad Ghassemi, Benjamin Moody, Peter Szolovits, Leo~Anthony Celi, and
  Roger~G Mark. 2016.
\newblock \href {https://doi.org/10.13026/C2XW26} {Mimic-iii, a freely
  accessible critical care database}.
\newblock \emph{Scientific data}, 3(1):1--9.

\bibitem[{Johnson et~al.(2020)Johnson, Mak, Barker, and
  Loessberg-Zahl}]{johnson-etal-2020-probing}
Devin Johnson, Denise Mak, Andrew Barker, and Lexi Loessberg-Zahl. 2020.
\newblock \href {https://doi.org/10.18653/v1/2020.blackboxnlp-1.18} {Probing
  for multilingual numerical understanding in transformer-based language
  models}.
\newblock In \emph{Proceedings of the Third BlackboxNLP Workshop on Analyzing
  and Interpreting Neural Networks for NLP}, pages 184--192, Online.
  Association for Computational Linguistics.

\bibitem[{Lan et~al.(2020)Lan, Chen, Goodman, Gimpel, Sharma, and
  Soricut}]{Lan2020ALBERT}
Zhenzhong Lan, Mingda Chen, Sebastian Goodman, Kevin Gimpel, Piyush Sharma, and
  Radu Soricut. 2020.
\newblock \href {https://openreview.net/forum?id=H1eA7AEtvS} {Albert: A lite
  bert for self-supervised learning of language representations}.
\newblock In \emph{International Conference on Learning Representations}.

\bibitem[{Lee et~al.(2020)Lee, Yoon, Kim, Kim, Kim, So, and
  Kang}]{lee2020biobert}
Jinhyuk Lee, Wonjin Yoon, Sungdong Kim, Donghyeon Kim, Sunkyu Kim, Chan~Ho So,
  and Jaewoo Kang. 2020.
\newblock \href {https://doi.org/10.1093/bioinformatics/btz682} {Biobert: a
  pre-trained biomedical language representation model for biomedical text
  mining}.
\newblock \emph{Bioinformatics}, 36(4):1234--1240.

\bibitem[{Lin et~al.(2020)Lin, Lee, Khanna, and Ren}]{lin-etal-2020-birds}
Bill~Yuchen Lin, Seyeon Lee, Rahul Khanna, and Xiang Ren. 2020.
\newblock \href {https://doi.org/10.18653/v1/2020.emnlp-main.557} {{B}irds have
  four legs?! {N}umer{S}ense: {P}robing {N}umerical {C}ommonsense {K}nowledge
  of {P}re-{T}rained {L}anguage {M}odels}.
\newblock In \emph{Proceedings of the 2020 Conference on Empirical Methods in
  Natural Language Processing (EMNLP)}, pages 6862--6868, Online. Association
  for Computational Linguistics.

\bibitem[{Naik et~al.(2019)Naik, Ravichander, Rose, and
  Hovy}]{naik-etal-2019-exploring}
Aakanksha Naik, Abhilasha Ravichander, Carolyn Rose, and Eduard Hovy. 2019.
\newblock \href {https://doi.org/10.18653/v1/P19-1329} {Exploring numeracy in
  word embeddings}.
\newblock In \emph{Proceedings of the 57th Annual Meeting of the Association
  for Computational Linguistics}, pages 3374--3380, Florence, Italy.
  Association for Computational Linguistics.

\bibitem[{Pal and Baral(2021)}]{pal-baral-2021-investigating-numeracy}
Kuntal~Kumar Pal and Chitta Baral. 2021.
\newblock \href {https://doi.org/10.18653/v1/2021.findings-emnlp.265}
  {Investigating numeracy learning ability of a text-to-text transfer model}.
\newblock In \emph{Findings of the Association for Computational Linguistics:
  EMNLP 2021}, pages 3095--3101, Punta Cana, Dominican Republic. Association
  for Computational Linguistics.

\bibitem[{Paszke et~al.(2019)Paszke, Gross, Massa, Lerer, Bradbury, Chanan,
  Killeen, Lin, Gimelshein, Antiga, Desmaison, Kopf, Yang, DeVito, Raison,
  Tejani, Chilamkurthy, Steiner, Fang, Bai, and Chintala}]{NEURIPS2019_9015}
Adam Paszke, Sam Gross, Francisco Massa, Adam Lerer, James Bradbury, Gregory
  Chanan, Trevor Killeen, Zeming Lin, Natalia Gimelshein, Luca Antiga, Alban
  Desmaison, Andreas Kopf, Edward Yang, Zachary DeVito, Martin Raison, Alykhan
  Tejani, Sasank Chilamkurthy, Benoit Steiner, Lu~Fang, Junjie Bai, and Soumith
  Chintala. 2019.
\newblock \href
  {http://papers.neurips.cc/paper/9015-pytorch-an-imperative-style-high-performance-deep-learning-library.pdf}
  {Pytorch: An imperative style, high-performance deep learning library}.
\newblock In H.~Wallach, H.~Larochelle, A.~Beygelzimer, F.~d\textquotesingle
  Alch\'{e}-Buc, E.~Fox, and R.~Garnett, editors, \emph{Advances in Neural
  Information Processing Systems 32}, pages 8024--8035. Curran Associates, Inc.

\bibitem[{Peng et~al.(2020)Peng, Chen, and Lu}]{peng-etal-2020-empirical}
Yifan Peng, Qingyu Chen, and Zhiyong Lu. 2020.
\newblock \href {https://doi.org/10.18653/v1/2020.bionlp-1.22} {An empirical
  study of multi-task learning on {BERT} for biomedical text mining}.
\newblock In \emph{Proceedings of the 19th SIGBioMed Workshop on Biomedical
  Language Processing}, pages 205--214, Online. Association for Computational
  Linguistics.

\bibitem[{Petroni et~al.(2019)Petroni, Rockt{\"a}schel, Riedel, Lewis, Bakhtin,
  Wu, and Miller}]{petroni-etal-2019-language}
Fabio Petroni, Tim Rockt{\"a}schel, Sebastian Riedel, Patrick Lewis, Anton
  Bakhtin, Yuxiang Wu, and Alexander Miller. 2019.
\newblock \href {https://doi.org/10.18653/v1/D19-1250} {Language models as
  knowledge bases?}
\newblock In \emph{Proceedings of the 2019 Conference on Empirical Methods in
  Natural Language Processing and the 9th International Joint Conference on
  Natural Language Processing (EMNLP-IJCNLP)}, pages 2463--2473, Hong Kong,
  China. Association for Computational Linguistics.

\bibitem[{Spithourakis and Riedel(2018)}]{spithourakis-riedel-2018-numeracy}
Georgios Spithourakis and Sebastian Riedel. 2018.
\newblock \href {https://doi.org/10.18653/v1/P18-1196} {Numeracy for language
  models: Evaluating and improving their ability to predict numbers}.
\newblock In \emph{Proceedings of the 56th Annual Meeting of the Association
  for Computational Linguistics (Volume 1: Long Papers)}, pages 2104--2115,
  Melbourne, Australia. Association for Computational Linguistics.

\bibitem[{Talmor et~al.(2020)Talmor, Elazar, Goldberg, and
  Berant}]{talmor-etal-2020-olmpics}
Alon Talmor, Yanai Elazar, Yoav Goldberg, and Jonathan Berant. 2020.
\newblock \href {https://doi.org/10.1162/tacl_a_00342} {o{LM}pics-on what
  language model pre-training captures}.
\newblock \emph{Transactions of the Association for Computational Linguistics},
  8:743--758.

\bibitem[{Wallace et~al.(2019)Wallace, Wang, Li, Singh, and
  Gardner}]{wallace-etal-2019-nlp}
Eric Wallace, Yizhong Wang, Sujian Li, Sameer Singh, and Matt Gardner. 2019.
\newblock \href {https://doi.org/10.18653/v1/D19-1534} {Do {NLP} models know
  numbers? probing numeracy in embeddings}.
\newblock In \emph{Proceedings of the 2019 Conference on Empirical Methods in
  Natural Language Processing and the 9th International Joint Conference on
  Natural Language Processing (EMNLP-IJCNLP)}, pages 5307--5315, Hong Kong,
  China. Association for Computational Linguistics.

\bibitem[{Wolf et~al.(2020)Wolf, Debut, Sanh, Chaumond, Delangue, Moi, Cistac,
  Rault, Louf, Funtowicz, Davison, Shleifer, von Platen, Ma, Jernite, Plu, Xu,
  Le~Scao, Gugger, Drame, Lhoest, and Rush}]{wolf-etal-2020-transformers}
Thomas Wolf, Lysandre Debut, Victor Sanh, Julien Chaumond, Clement Delangue,
  Anthony Moi, Pierric Cistac, Tim Rault, Remi Louf, Morgan Funtowicz, Joe
  Davison, Sam Shleifer, Patrick von Platen, Clara Ma, Yacine Jernite, Julien
  Plu, Canwen Xu, Teven Le~Scao, Sylvain Gugger, Mariama Drame, Quentin Lhoest,
  and Alexander Rush. 2020.
\newblock \href {https://doi.org/10.18653/v1/2020.emnlp-demos.6} {Transformers:
  State-of-the-art natural language processing}.
\newblock In \emph{Proceedings of the 2020 Conference on Empirical Methods in
  Natural Language Processing: System Demonstrations}, pages 38--45, Online.
  Association for Computational Linguistics.

\bibitem[{Wu et~al.(2021)Wu, Zhang, Wei, and Huang}]{wu-etal-2021-math}
Qinzhuo Wu, Qi~Zhang, Zhongyu Wei, and Xuanjing Huang. 2021.
\newblock \href {https://doi.org/10.18653/v1/2021.acl-long.455} {Math word
  problem solving with explicit numerical values}.
\newblock In \emph{Proceedings of the 59th Annual Meeting of the Association
  for Computational Linguistics and the 11th International Joint Conference on
  Natural Language Processing (Volume 1: Long Papers)}, pages 5859--5869,
  Online. Association for Computational Linguistics.

\bibitem[{Yang et~al.(2021)Yang, Chen, Chen, and Cer}]{yang2021nt5}
Peng~Jian Yang, Ying~Ting Chen, Yuechan Chen, and Daniel Cer. 2021.
\newblock \href {http://arxiv.org/abs/2104.07307} {Nt5?! training {T5} to
  perform numerical reasoning}.
\newblock \emph{arXiv preprint arXiv:2104.07307}.

\bibitem[{Yuan et~al.(2021)Yuan, Neubig, and Liu}]{yuan-neurips-2021-bartscore}
Weizhe Yuan, Graham Neubig, and Pengfei Liu. 2021.
\newblock \href
  {https://proceedings.neurips.cc/paper/2021/file/e4d2b6e6fdeca3e60e0f1a62fee3d9dd-Paper.pdf}
  {Bartscore: Evaluating generated text as text generation}.
\newblock In \emph{Advances in Neural Information Processing Systems},
  volume~34, pages 27263--27277. Curran Associates, Inc.

\bibitem[{Zhang et~al.(2020)Zhang, Ramachandran, Tenney, Elazar, and
  Roth}]{zhang-etal-2020-language-embeddings}
Xikun Zhang, Deepak Ramachandran, Ian Tenney, Yanai Elazar, and Dan Roth. 2020.
\newblock \href {https://doi.org/10.18653/v1/2020.findings-emnlp.439} {Do
  language embeddings capture scales?}
\newblock In \emph{Findings of the Association for Computational Linguistics:
  EMNLP 2020}, pages 4889--4896, Online. Association for Computational
  Linguistics.

\end{thebibliography}
